\def\year{2019}\relax
\documentclass[letterpaper]{article} 
\usepackage{aaai19}  
\usepackage{times}  
\usepackage{helvet}  
\usepackage{courier}  
\usepackage{url}  
\usepackage{graphicx}  
\usepackage[utf8]{inputenc} 
\usepackage[T1]{fontenc}    
\usepackage{booktabs}       
\usepackage{amsfonts}       
\usepackage{nicefrac}       
\usepackage{microtype}      

\usepackage{graphics}
\usepackage{graphicx}
\usepackage{algorithm}
\usepackage{algorithmicx}
\usepackage[noend]{algpseudocode}
\usepackage{amssymb}
\usepackage{amsmath }
\usepackage{subfigure}

\algnewcommand{\LineComment}[1]{\State \(//\) #1}

\newcommand{\textcite}[1]{\cite{\#1}}

\newlength\myindent
\begin{document}
%
\title{Biologically Motivated Algorithms for Propagating Local Target Representations}
\author{Alexander G. Ororbia\thanks{ \hspace{0.1cm} Both authors contributed equally.}\\
Rochester Institute of Technology\\
102 Lomb Memorial Drive, Rochester NY, USA 14623\\
{\tt ago@cs.rit.edu}\\
\And 
Ankur Mali\footnotemark[1] \\
Penn State University\\
Old Main, State College, PA 16801\\
{\tt aam35@ist.psu.edu}
}
\maketitle
\begin{abstract}
Finding biologically plausible alternatives to back-propagation of errors is a fundamentally important challenge in artificial neural network research. In this paper, we propose a learning algorithm called error-driven Local Representation Alignment (LRA-E), which has strong connections to predictive coding, a theory that offers a mechanistic way of describing neurocomputational machinery. In addition, we propose an improved variant of Difference Target Propagation, another procedure that comes from the same family of algorithms as LRA-E. We compare our procedures to several other biologically-motivated algorithms, including two feedback alignment algorithms and Equilibrium Propagation. In two benchmarks, we find that both of our proposed algorithms yield stable performance and strong generalization compared to other competing back-propagation alternatives when training deeper, highly nonlinear networks, with LRA-E performing the best overall.
\end{abstract}

\noindent Behind the modern achievements in artificial neural network research is back-propagation of errors \cite{rumelhart1988backprop} (or ``backprop''), the key training algorithm used in computing updates for the parameters that define the computational architectures applied to problems ranging from computer vision to speech recognition. However, though neural architectures are inspired by our current understanding of the human brain, the connections to the actual mechanisms of systems of natural neurons are often very loose, at best. More importantly, backprop faces some of the strongest neuro-biological criticisms, argued to be a highly implausible way in which learning occurs in the human brain.

Among the many problems with back-propagation, some of the most prominent are:
1) the ``weight transport problem'', where the feedback weights that carry back error signals must be the transposes of the feedforward weights, 2) forward and backward propagation utilize different computations, and 3) the error gradients are stored separately from the activations.
These problems, as originally argued in \cite{ororbia2017learning,ororbia2018conducting}, largely center around one critical component of backprop--the global feedback pathway needed for transporting error derivatives across the system. This pathway is necessary given the design of modern supervised learning systems--a loss measures error between a model's output units and a target, e.g., class label, and the global pathway relates how the internal processing elements affect this error.
When considering modern theories of the brain \cite{grossberg1982does,rao1999predictive,huang2011predictive}, which posit that local computations occur at multiple levels of a somewhat hierarchical structure, this global pathway should not be necessary to learn effectively. Furthermore, this pathway results makes training very deep networks difficult--due to many multiplications that underly traversing along the global feedback pathway, error gradients explode/vanish \cite{pascanu2013difficulty}. To fix this, gradients are kept within reasonable magnitudes by requiring layers to behave sufficiently linearly.
However, this remedy creates other highly undesirable side-effects, e.g., adversarial samples \cite{ororbia2017unifying}, and prevents usage of neuro-biological mechanisms such as lateral competition and discrete-valued/stochastic activations (since the pathway requires precise knowledge of function derivatives~\cite{bengio2015towards}).

If we remove this global feedback pathway, we create a new problem--what are the learning signals for the hidden processing elements? This problem is one of the main concerns of the recently introduced \emph{Discrepancy Reduction} family of learning algorithms \cite{ororbia2017learning}. 
In this paper, we will develop two learning algorithms within this family--error-driven Local Representation Alignment and adaptive noise Difference Target Propagation. In experiments on two classification benchmarks, we will show that these two algorithms generalize better than a variety of other biologically motivated learning approaches, all without employing the global feedback pathway required by back-propagation.

\section{Coordinated Local Learning Algorithms} 
\label{sec:lra_dr}
Algorithms within the Discrepancy Reduction \cite{ororbia2017learning} family offer computational mechanisms for two key steps when learning from patterns. These steps include:
\begin{enumerate}
	\item Search for latent representations that better explain the input/output, also known as target representations. This creates the need for local (higher-level) objectives that will guide current latent representations towards better ones.
    \item Reduce, as much as possible, the mismatch between a model's currently ``guessed'' representations and target representations. The sum of the internal, local losses is also defined as the total discrepancy in a system, and can also be thought of as a sort of pseudo-energy function.
\end{enumerate}
This general process forms the basis of what we call \emph{coordinated local learning rules}. Computing targets with these kinds of rules should not require an actual pathway, as in back-propagation, and instead make use of top-down and bottom-up signals to generate targets. This idea is particularly motivated by the theory of predictive coding \cite{panichello2013predictive} (which started to impact modern machine learning applications \cite{li2018predictive}), which claims that the brain is in a continuous process of creating and updating hypotheses (using error information) to predict the sensory input.
This paper will explore two ways in which this hypothesis updating (in the form of local target creation) might happen: 1) through error-correction in Local Representation Alignment (LRA-E), and 2) through repeated encoding and decoding as in Difference Target Propagation (DTP).
%
It should also be noted that one is not restricted to only using neural building blocks--LRA-E could be used to train stacked Gradient Boosted Decision Trees (GBDTs), which would be faster than in \cite{feng2018multi}, which employed a form of target propagation to calculate local updates.

The idea of learning locally, in general, is slowly becoming prominent in the training of artificial neural networks, with recent proposals including decoupled neural interfaces \cite{jaderberg2016decoupled} and kickback \cite{balduzzi2015kickback} (which was derived specifically for regression problems). Furthermore, \cite{whittington2017equivalence} demonstrated that neural models using simple local Hebbian updates (within a predictive coding framework) could efficiently conduct supervised learning. Far earlier approaches that employed local learning included the layer-wise training procedures that were once used to build models for unsupervised learning \cite{bengio_greedy_2007}, supervised learning \cite{lee_deeply-supervised_2014}, and semi-supervised learning \cite{ororbia_deep_hybrid_2015a,ororbia2015online}.  The key problem with these older algorithms is that they were greedy--a model was built from the bottom-up, freezing lower-level parameters as higher-level feature detectors were learnt. 

Another important idea underlying algorithms such as LRA and DTP is that learning is possible with asymmetry--which directly resolves the weight-transport problem \cite{grossberg_resonance_1987,liao2016important}, another strong neuro-biological criticism of backprop. This is even possible, surprisingly, if the feedback weights are random and fixed, which is at the core of two algorithms we will also compare to--Random Feedback Alignment (RFA) \cite{lillicrap2016random} and Direct Feedback Alignment (DFA) \cite{nokland2016direct}. RFA replaces the transpose of the feedforward weights in backprop with a similarly-shaped random matrix while DFA directly connects the output layer's pre-activation derivative to each layer's post-activation. It was shown in \cite{ororbia2017learning,ororbia2018conducting} that these feedback loops would be better suited in generating target representations.

\subsection{Local Representation Alignment}
\label{lra}
To concretely describe how LRA is practically implemented, we will specify how LRA is applied to a 3-layer feedforward network, or multilayer perceptron (MLP). Note that LRA generalizes to models with more layers ($L \geq 3$). 

The pre-activities of the MLP at layer $\ell$ are denoted as $\mathbf{h}^\ell$ while the post-activities, or the values output by the non-linearity $\phi_\ell(\cdot)$, are denoted as $\mathbf{z}^\ell$. The target variable used to correct the output units ($\mathbf{z}^L$) is denoted as $\mathbf{y}^L_z$ ($\mathbf{y}^L_z = \mathbf{y}$, or $\mathbf{y}^L_z = \mathbf{x}$ if we are learning auto-associative functions). Connecting one layer of neurons $\mathbf{z}^{\ell-1}$, with pre-activities $\mathbf{h}^{\ell-1}$, to another layer $\mathbf{z}^{\ell}$, with pre-activities $\mathbf{h}^{\ell}$, are synaptic weights $W_\ell$. The  propagation equations for computing pre-activtion and post-activation values for layer $\ell$ are:
\begin{align}
\mathbf{h}^\ell &= W_\ell \mathbf{z}^{\ell-1}, \quad \mathbf{z}^\ell = \phi_\ell(\mathbf{h}^\ell) \label{eqn:fprop}
\end{align}
Before computing targets or updates, we first must define the set of local losses, one per layer of neurons except for the input neurons, that constitute the measure of total discrepancy inside the MLP, $\{ \mathcal{L}_1(\mathbf{y}^1_z,\mathbf{z}^1), \mathcal{L}_2(\mathbf{y}^2_z,\mathbf{z}^2), \mathcal{L}_3(\mathbf{y}^3_z,\mathbf{z}^3) \}$. With losses defined, we can then explicitly formulate the error units $\mathbf{e}_\ell$ for each layer as well, since any given layer's error units correspond to the first derivative of that layer's loss with respect to that layer's post-activation values. For the MLP's output layer, we could assume a categorical distribution, which is appropriate for 1-of-$k$ classification tasks, and use the following negative log likelihood loss:
\begin{align}
\mathcal{L}_\ell(\mathbf{y}^\ell_z, \mathbf{z}^\ell) &= -\frac{1}{2} \sum^{|\mathbf{z}|}_{i=1} \mathbf{y}^\ell_z[i] \log \mathbf{z}^\ell[i]  \mbox{,} \nonumber \\  
\mathbf{e}_\ell &= \mathbf{e}_\ell(\mathbf{y}^\ell_z, \mathbf{z}^\ell) = \frac{-\mathbf{y}^\ell_z}{\mathbf{z}^\ell} \mbox{,} \label{cat_loss}
\end{align}
where the loss is computed over all dimensions $|\mathbf{z}|$ of the vector $\mathbf{z}$ (where a dimension is indexed/accessed by integer $i$). Note that for this loss function, we assume that $\mathbf{z}$ is a vector of probabilities computed by using the softmax function as the output nonlinearity, $\mathbf{z}^3 = \frac{exp(\mathbf{h}^3)}{\sum_i exp(\mathbf{h}^3_i)}$. For the hidden layers, we can choose between a wider variety of loss functions, and in this paper, we experimented with assuming either a Gaussian or Cauchy distribution over the hidden units. For the Gaussian distribution (or L2 norm), we have the following:
\begin{align}
\mathcal{L}_\ell(\mathbf{z}, \mathbf{y}) = \frac{1}{(2 \sigma^2)} \sum^{|\mathbf{z}|}_{i=1} (\mathbf{y}_i - \mathbf{z}_i)^2 \nonumber \\
\mathbf{e}_\ell = \mathbf{e}_\ell(\mathbf{y}^\ell_z, \mathbf{z}^\ell) = \frac{-(\mathbf{y}^\ell_z - \mathbf{z}^\ell)}{\sigma^2} \label{gaussian_loss}
\end{align}
where $\sigma^2$ represents fixed scalar variance (we set $\sigma^2 = 1/2$). For the Cauchy distribution (or log-penalty), we obtain:
\begin{align}
\mathcal{L}_\ell(\mathbf{z}, \mathbf{y}) = \sum^{|\mathbf{z}|}_{i=1} \log(1 + (\mathbf{y}_i - \mathbf{z}_i)^2) \nonumber \\
\mathbf{e}_\ell = \mathbf{e}_\ell(\mathbf{y}^\ell_z, \mathbf{z}^\ell) = \frac{-2 (\mathbf{y}^\ell_z - \mathbf{z}^\ell)}{(1 + (\mathbf{y}^\ell_z - \mathbf{z}^\ell)^2)} \mbox{.} \label{cauchy_loss}
\end{align}
For the activation function used in calculating the hidden post-activities, we use the hyperbolic tangent, or $\phi_\ell(v) = \frac{exp(2v) - 1}{exp(2v) + 1}$. Using the Cauchy distribution proved particularly useful in our experiments because it encourages sparse representations and aligns nicely with the biological considerations of sparse coding \cite{olshausen1997sparse} and predictive sparse decomposition \cite{kavukcuoglu2010fast} as well as the lateral competition \cite{rao1999predictive} that naturally occurs in groups of neural processing elements. 
These are relatively simple local losses for measuring the agreement between representations and targets and future work should entail developing even better metrics.

\begin{algorithm}[H]
\caption{LRA-E: Target and update computations.}
\label{alg:lra_e}
\begin{algorithmic}
   \LineComment Procedure for computing error units \& targets
   \State {\bfseries Input:} sample $(\mathbf{y},\mathbf{x})$ and $ \Theta = \{ W_1, W_2, W_3, E_2, E_3 \}$
   \Function{ComputeTargets}{$(\mathbf{y},\mathbf{x}), \Theta$}
   		\LineComment Run feedforward weights to get activities
    	\State $\mathbf{h}^1 = W_1 \mathbf{z}^0$, $\mathbf{z}^1 = \phi_1(\mathbf{h}^1) $
    	\State $\mathbf{h}^2 = W_2 \mathbf{z}^1$, $\mathbf{z}^2 = \phi_2(\mathbf{h}^2) $
        \State $\mathbf{h}^3 = W_3 \mathbf{z}^2$, $\mathbf{z}^3 = \phi_3(\mathbf{h}^3) $
        \State $\mathbf{y}^3_z \Leftarrow \mathbf{y}$
        \State $\mathbf{e}_3 = \frac{-\mathbf{y}^3_z}{\mathbf{z}^3} $, $\mathbf{y}^2_z \leftarrow \phi_2 \Big( \mathbf{h}^2 - \beta ( E_3 \mathbf{e}_3 ) \Big)$
        \State $\mathbf{e}_2 = -2 ( \mathbf{y}^2_z - \mathbf{z}^2 ) $
        \State $\mathbf{y}^1_z \leftarrow \phi_1 \Big( \mathbf{h}^1 - \beta ( E_2 \mathbf{e}_2 ) \Big)$
        \State $\mathbf{e}_1 = -2 ( \mathbf{y}^1_z - \mathbf{z}^1 ) $
        \State $\Lambda = (\mathbf{z}^3,\mathbf{z}^2,\mathbf{z}^1,\mathbf{h}^3,\mathbf{h}^2,\mathbf{h}^1,\mathbf{e}^3,\mathbf{e}^2,\mathbf{e}^1)$
        \State \textbf{Return} $\Lambda$
    \EndFunction
\end{algorithmic}
\begin{algorithmic}
   \LineComment Procedure(s) for computing weight updates
   \State {\bfseries Input:} sample $(\mathbf{y},\mathbf{x})$  and calculations $\Lambda$
   \Function{CalcUpdates-V1}{$(\mathbf{y},\mathbf{x}), \Theta, \Lambda$}
   		\State $\Delta W_3 = (\mathbf{e}_3 \otimes  \phi^\prime_3(\mathbf{h}_3) ) (\mathbf{z}_2)^T$
        \State $\Delta W_2 = (\mathbf{e}_2 \otimes  \phi^\prime_2(\mathbf{h}_2) ) (\mathbf{z}_1)^T$
        \State $\Delta W_1 = (\mathbf{e}_1 \otimes  \phi^\prime_1(\mathbf{h}_1) ) (\mathbf{x})^T$
        \State $ \Delta E_3 = -\gamma (\Delta W_3)^T $
        \State $ \Delta E_2 = -\gamma (\Delta W_2)^T $
        \State \textbf{Return} $ \big( \Delta W_3,\Delta W_2,\Delta W_1,\Delta E_3,\Delta E_2 \big) $
    \EndFunction
   \Function{CalcUpdates-V2}{$(\mathbf{y},\mathbf{x}), \Theta, \Lambda$}
        \State $\Delta W_3 = \mathbf{e}_3 (\mathbf{z}_2)^T$
        \State $\Delta W_2 = \mathbf{e}_2 (\mathbf{z}_1)^T$
        \State $\Delta W_1 = \mathbf{e}_1 (\mathbf{x})^T$
        \State $ \Delta E_3 = -\gamma (\Delta W_3)^T $
        \State $ \Delta E_2 = -\gamma (\Delta W_2)^T $
        \State \textbf{Return} $ \big( \Delta W_3,\Delta W_2,\Delta W_1,\Delta E_3,\Delta E_2 \big) $
    \EndFunction
\end{algorithmic}
\end{algorithm}

With local losses specified and error units implemented, all that remains is to define how targets are computed and what the parameter updates will be. At any given layer $\mathbf{z}^\ell$, starting at the output units (in our example, $\mathbf{z}^3$), we calculate the target for the layer below $\mathbf{z}^{\ell-1}$ by multiplying the error unit values at $\ell$ by a set of synaptic error weights $E_\ell$. This projected displacement, weighted by the modulation factor $\beta$,\footnote{In this paper, $\beta = 0.1$, found with only minor prelim. tuning.} is then subtracted from the initially found pre-activation of the layer below $\mathbf{h}^{\ell-1}$. This updated pre-activity is then run through the appropriate nonlinearity to calculate the final target $\mathbf{y}^{\ell-1}_z$. This computation amounts to:
\begin{align}
\mathbf{e}_\ell &= -2 ( \mathbf{y}^\ell_z - \mathbf{z}^\ell ) \mbox{,} \quad \Delta \mathbf{h}^{\ell-1} = E_\ell \mathbf{e}_\ell \\
\mathbf{y}^{\ell-1}_z &\leftarrow \phi_{\ell-1} \Big( \mathbf{h}^{\ell-1} - \beta ( \Delta \mathbf{h}^{\ell-1} ) \Big) \mbox{.} \label{eqn:displace}
\end{align}
Once the targets for each layer have been found, we can then use the local loss $\mathcal{L}^{\ell}(\mathbf{y}^{\ell}_z,\mathbf{z}^{\ell})$ to compute updates to the weights $W_{\ell}$ and its corresponding error weights $E_{\ell}.$\footnote{Except for $W_1$, which has no corresponding error weights $E_1$.} The update calculation for parameters at layer $\ell$ would be:
\begin{align}
  \Delta W_{\ell} = (\mathbf{e}_{\ell} \otimes  \phi^\prime_\ell(\mathbf{h}_\ell) )(\mathbf{z}_{\ell-1})^T \mbox{, } \Delta E_{\ell} &= -\gamma (\Delta W_\ell)^T, \\
  \mbox{or, } \quad  \Delta W_{\ell} = \mathbf{e}_{\ell} (\mathbf{z}_{\ell-1})^T \mbox{, } \Delta E_{\ell} &= -\gamma (\Delta W_\ell)^T
\end{align}
where $\otimes$ indicates the Hadamard product and $\gamma$ is a decay factor (a value that we found should be set to less than $1.0$) meant to ensure that the error weights change more slowly than the forward weights. An attractive property of LRA is that the derivatives of the pointwise activation functions can be dropped, yielding the second variation of the update rule, as long as the activation function is monotonically non-decreasing in its input (for stochastic activation functions, the output distribution for a larger input should stochastically dominate the output distribution for a smaller input). This is also satisfying from a biological perspective since it is unlikely that neurons would utilize point-wise activation derivatives in computing updates \cite{hinton1988learning}.
The update for error weights is simply proportional to the negative transpose of the update computed for the matching forward weights, which is a computationally fast and cheap rule we propose inspired by \cite{rao1997dynamic}. 

In Algorithm \ref{alg:lra_e}, the equations in this section are combined to create the full procedure for training a 3-layer MLP (using either $\Call{CALCUPDATES-V1}{\cdot}$ or $\Call{CALCUPDATES-V2}{\cdot}$ to compute weight updates), assuming Gaussian local losses and their respective error units. The model is defined by $\Theta = \{ W_1, W_2, W_3, E_2, E_3\}$ (biases $\mathbf{c}_\ell$ omitted for clarity). We will refer to Algorithm \ref{alg:lra_e} as \emph{LRA-E} (which easily extends to $L > 3$).

\begin{figure*}
\centering     
\subfigure[Gradient angle (degrees) compared to BP.]{\label{fig:angles}\includegraphics[width=57mm]{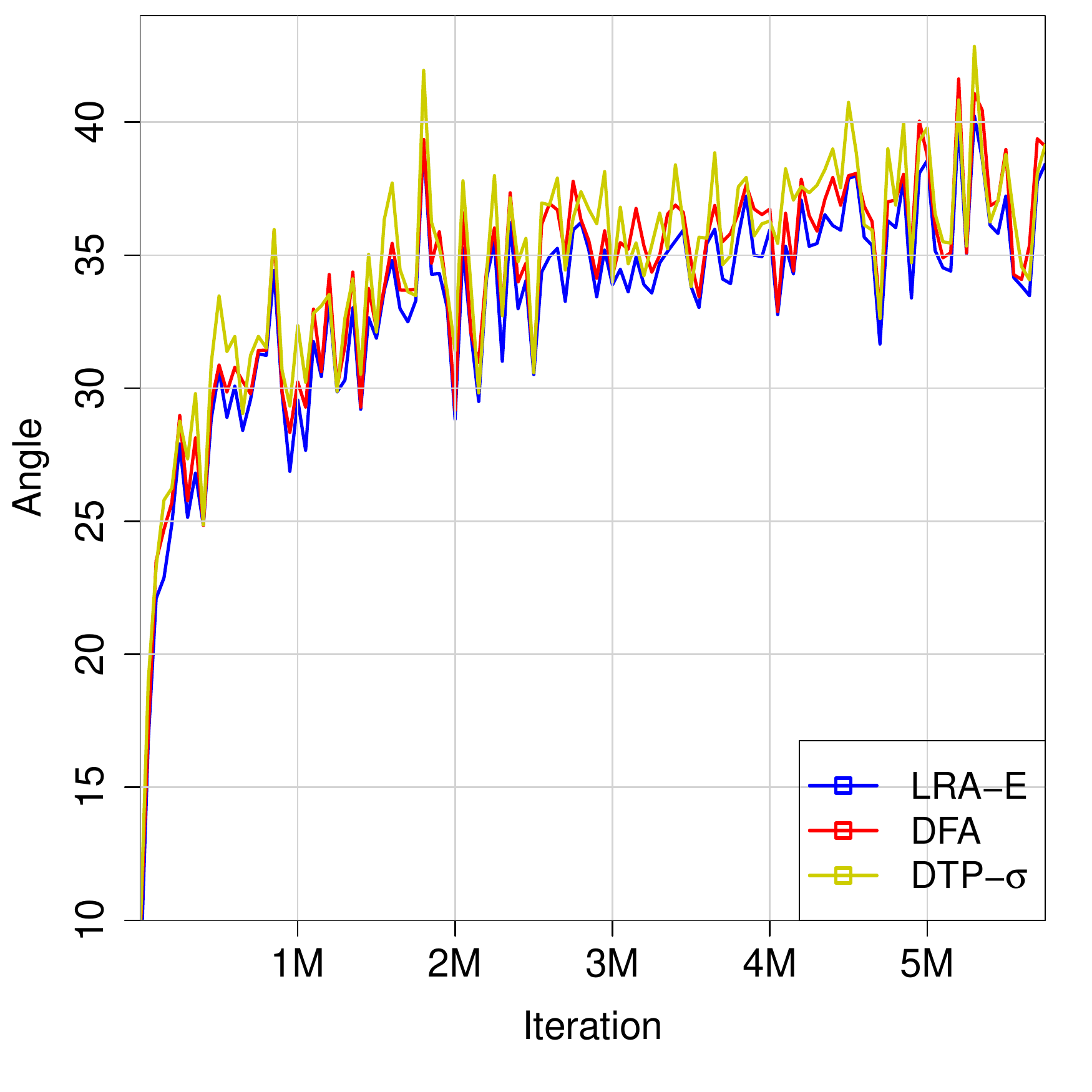}
}
\hspace{0.15cm}
\subfigure[Total discrepancy \& output NLL.]{\label{fig:discrep}\includegraphics[width=57mm]{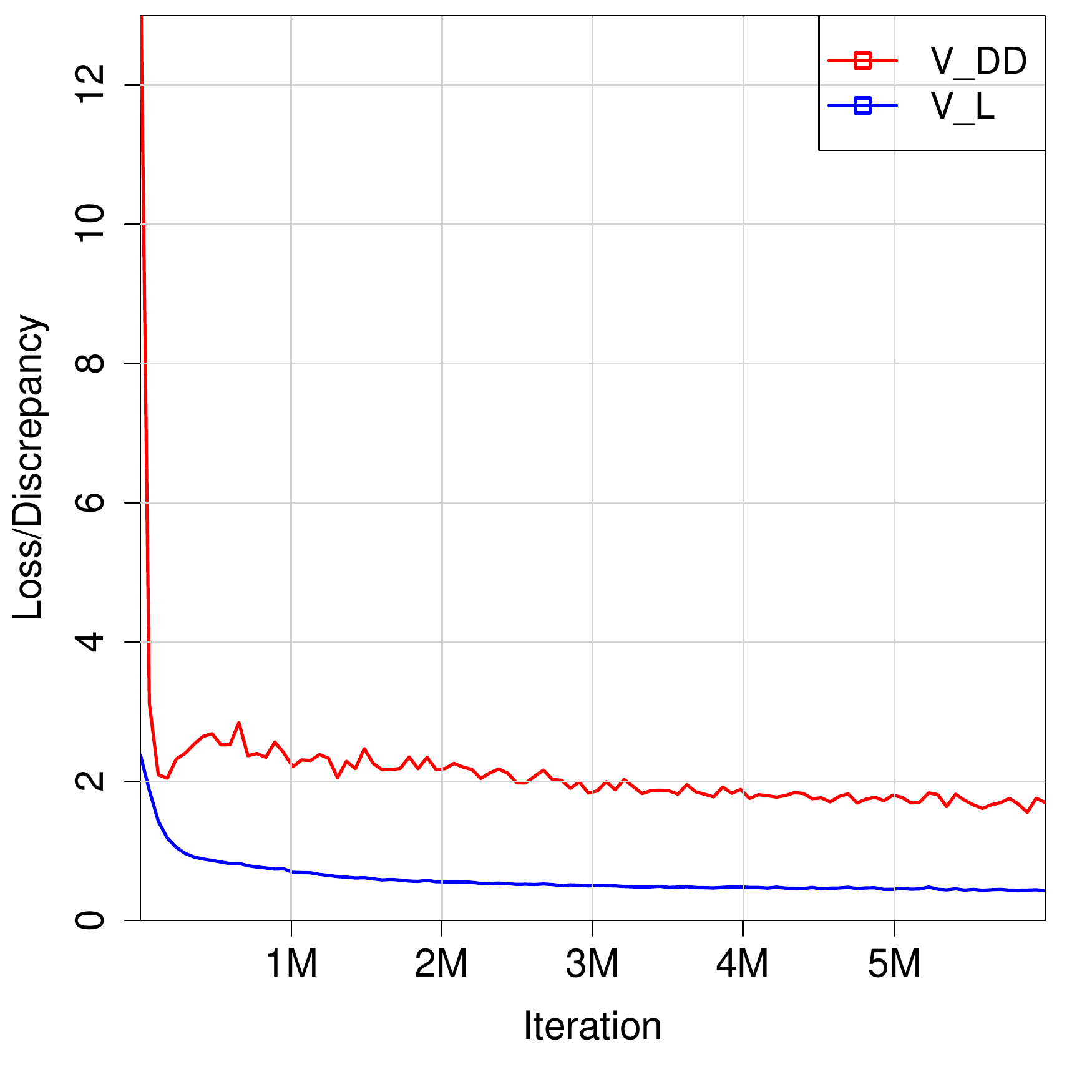}
}
\vspace{-0.3cm}
\caption{In Figure \ref{fig:angles}, we compare the updates calculated by LRA-E, DFA, and DTP-$\sigma$ against backprop (BP). In Figure \ref{fig:discrep}, we show how total discrepancy for an LRA-trained MLP evolves during training on Fashion MNIST, alongside the output loss.}
\label{fig:lra_algos}
\end{figure*}

In Figure \ref{fig:angles}, we compare the updates calculated by \emph{LRA-E} (as well as DFA and our proposed DTP-$\sigma$, described later) with those given by back-propagation, after each mini-batch, by plotting the angles over the first 20 epochs of learning for a 3-layer MLP (256 units per layer) trained with stochastic gradient descent (SGD) with mini-batches of 50 image samples using a categorical output loss and Gaussian local losses.
As long as the angle of the updates computed from LRA are within 90 degrees of the updates obtained by back-propagation, LRA will move parameters towards the same general direction as back-propagation (which greedily points in the direction of steepest descent) and will still find good local optima. In Figure \ref{fig:angles}, this does indeed appear to be the case for the MLP example. The angle, fortunately, while certainly non-zero, never deviates too far from the direction pointed by back-propagation and remains relatively stable throughout the learning process. (Observe that DFA and DTP~-$\sigma$ have, interestingly enough, update angles that are quite similar to LRA-E.)
Alongside Figure~\ref{fig:angles}, in Figure~\ref{fig:discrep}, we plot our neural model's total internal discrepancy, $D(\mathbf{y},\mathbf{x})$ (or \emph{V\_DD}), which is a simple linear combination of all of the internal local losses for a given data point. 
Observe that while the (validation) output loss (\emph{V\_L}) continually decreases, \emph{V\_DD} does not always appear to do so. 
We conjecture that this ``bump'', which appears at the start of learning, is the result of the evolution of LRA-E's error weights, which are used to directly control the target generation process. So even though backprop and LRA-E might start down the same path in error space (or on the loss surface), as indicated by the initially low angle between updates, this trajectory is not ideal for LRA's units/targets. This means that error weights will change more rapidly at training's start, resulting in targets that vary quite a bit (raising internal loss values). However, once the error weights start to converge to an approximate transpose of the feedforward weights, the process of correction becomes easier and \emph{V\_DD} desirably declines. 


\subsection{Improving Difference Target Propagation}
\label{sec:dtp_sigma}
As mentioned earlier, Difference Target Propagation (DTP) (and also, less directly, recirculation \cite{hinton1988learning,o1996biologically}), like LRA-E, also falls under the same family of algorithms concerned with minimizing internal discrepancy, as shown in \cite{ororbia2017learning,ororbia2018conducting}. However, DTP takes a very different approach to computing alignment targets--instead of transmitting messages through error units and error feedback weights as in LRA-E, DTP employs feedback weights to learn the inverse of the mapping created by the feedforward weights. 
However, \cite{ororbia2018conducting} showed that DTP struggles to assign good local targets as the network becomes deeper, i.e., more highly nonlinear, facing an initially promising albeit brief phase in learning where generalization error decreases (within the first few epochs) before ultimately collapsing (unless very specific initializations are used). One potential cause of this failure could be the lack of a strong enough mechanism to globally coordinate the local learning problems created by the encoder-decoder pairs that underlie the system. In particular, we hypothesize that this problem might be coming from the noise injection scheme, which is local and fixed, offering no adaptation to each specific layer and making some of the layerwise optimization problems more difficult than necessary. Here, we will aim to remove this potential cause through an adaptive layerwise corruption scheme.

Assuming we have a target calculated from above $\mathbf{y}^{\ell}_z$, we consider the forward weights $W_\ell$ connecting the layer $\mathbf{z}^{\ell-1}$ to layer $\mathbf{z}^{\ell}$ and the decoding weights $E_\ell$ that define the inverse mapping between the two. The first forward propagation step is the same as in Equation \ref{eqn:fprop}.
In contrast to LRA-E's error-driven way of computing targets, we consider each pair of neuronal layers, $(\mathbf{z}^{\ell}, \mathbf{z}^{\ell-1})$, as forming a particular type of encoding/decoding cycle that will be used in computing layerwise targets. To calculate the target $\mathbf{y}^{\ell-1}_z$, we update the original post-activation $\mathbf{z}^{\ell-1}$ using the linear combination of two applications of the decoding weights as follows:
\begin{align}
	\mathbf{y}^{\ell-1}_z = \mathbf{z}^{\ell-1} - \big ( \phi_{\ell-1}( E_\ell \mathbf{z}^\ell ) + \phi_{\ell-1}( E_\ell \mathbf{y}^\ell_z ) \big )
\end{align}
where we see that we decode two times, one from the original post-activation calculated from the feedforward pass of the MLP and another from the target value generated from the encoding/decoding process from the layer pair above, e.g. $(\mathbf{z}^{\ell+1}, \mathbf{z}^{\ell})$. This will serve as the target when training the forward weights for the layer below $W_{\ell-1}$. To train the inverse-mapping weights $E_{\ell}$, as required by the original version of DTP, zero-mean Gaussian noise, $\epsilon \sim \mathcal{N}(0,\sigma^2)$ with fixed standard deviation $\sigma$, is injected into $\mathbf{z}^{\ell-1}$ followed by re-running the encoder and the decoder on this newly corrupted activation vector. Formally, this is defined as:
\begin{align}
	\mathbf{\widehat{y}}^{\ell-1}_z = \mathbf{z}^{\ell-1} + \epsilon \mbox{,} \quad \mathbf{\widehat{z}}^{\ell-1} = \phi_{\ell-1}(E_{\ell}\phi_{\ell}(W_\ell \mathbf{\widehat{y}}^{\ell-1}_z))
\end{align}
This process we will refer to as \emph{DTP}. 
In our proposed, improved variation of DTP, or \emph{DTP}-$\sigma$, we will take an ``adaptive'' approach to the noise injection process $\epsilon$. To develop our adaptive noise scheme, we have taken some insights from studies of biological neuron systems, which show there are varying levels of signal corruption in different neuronal layers/groups \cite{adrian,TOMKO1974405,TOLHURST1983775,pmid9570816}. It has been argued that this noise variability enhances neurons' overall ability to detect and transmit signals across a system \cite{shu,Kruglikov,pmid9570816} and, furthermore, that the presence of this noise yields more robust representations \cite{Cordo,pmid9570816,faizal}. There also is biological evidence demonstrating that an increase in the noise level across successive groups of neurons is thought to help in local neural computation \cite{pmid9570816,Sarpeshkar,Laughlin}.

In light of this, the standard deviation $\sigma$ of the noise process should be a function of the noise across layers, and an interesting way in which we implemented this was to make $\sigma_\ell$ (the standard deviation of the noise injection at layer $\ell$) a function of the local loss measurements. At the top layer, we can set $\sigma_L = \alpha$ (a small, fixed value such as $\alpha = 0.01$ worked well in our experiments). The standard deviation for the layers below would be a function of where the noise process is within the network, indexed by $\ell$. This means that:
\begin{align}
	\sigma_{\ell} = \sigma_{\ell+1} - \mathcal{L}_{\ell}(\mathbf{y}^{\ell-1}_z, \mathbf{z}^{\ell-1})
\end{align}
noting that the local loss chosen for DTP is a Gaussian loss (but with the input arguments flipped--the target value is now the corrupted initial encoding and the prediction is the clean, original encoding, or $\mathcal{L}_{\ell}(\mathbf{z} = \mathbf{y}^{\ell-1}_z, \mathbf{y} = \mathbf{z}^{\ell-1})$).

The updates to the weights are calculated by differentiating each local loss with respect to the appropriate encoder weights, or $\Delta W_{\ell-1} = \frac{\partial \mathcal{L}( \mathbf{z}^{\ell-1}, \mathbf{y}^{\ell-1}_z )}{\partial W_{\ell-1}}$, or with respect to the decoder synaptic weights $\Delta E_{\ell} = \frac{\partial \mathcal{L}( \mathbf{\widehat{z}}^{\ell}, \mathbf{\widehat{y}}^{\ell}_z )}{\partial E_{\ell}}$. Note that the order of the input arguments to each loss function for these two partial derivatives is important for obtaining the correct sign to multiply the gradients by, and furthermore staying aligned with the original formulation of DTP \cite{lee2015difference}, .

As we will see in our experimental results, \emph{DTP}-$\sigma$ is a much more stable learning algorithm (especially with respect to the original DTP), especially when training deeper/wider networks. \emph{DTP}-$\sigma$ benefits from a stronger form of overall coordination among its internal encoding/decoding sub-problems through the pair-wise comparison of local loss values that drive the hidden layer corruption. 

\subsection{A Comment on the Efficiency of LRA-E and DTP}
Note that \emph{LRA-E}, while a bit slower than backprop per update (given its use of the error weights to generate hidden layer targets), is much faster than \emph{DTP} and \emph{DTP}-$\sigma$. Specifically, if we focus on matrix multiplications used to find targets, which make up the bulk of the computation underlying both processes, \emph{LRA-E} only requires $2 (L - 1)$ matrix multiplications while \emph{DTP} and \emph{DTP}-$\sigma$ require $4 (L-3) + L$ multiplications. In particular, DTP has a very expensive target generation phase, requiring 2 applications of the encoder parameters (1 of these is from the network's initial feedfoward pass) and 3 applications of the decoder parameters to create targets to train the forward weights and inverse-mapping weights. 

\section{Experimental Results}
\label{sec:results}
In this section, we present experimental results of training MLPs using a variety of learning algorithms.

\textbf{MNIST:} This dataset \footnote{Available at the URL:  http://yann.lecun.com/exdb/mnist/.} contains $28\times28$ images with gray-scale pixel feature values in the range of $[0,255]$. The only preprocessing applied to this data is to normalize the pixel values to the range of $[0,1]$ by dividing them by 255.

\textbf{Fashion MNIST:} This database \cite{xiao2017fashion} contains $28x28$ grey-scale images of clothing items, meant to serve as a much more difficult drop-in replacement for MNIST itself. Training contains 60000 samples and testing contains 10000, each image is associated with one of 10 classes. We create a validation set of 2000 samples from the training split. Preprocessing was the same as on MNIST.

For both datasets and all models, over 100 epochs, we calculate updates over mini-batches of 50 samples. Furthermore, we do not regularize parameters any further, e.g., drop-out \cite{srivastava2014dropout} or weight penalties.
All feedfoward architectures for all experiments were of either $3$, $5$, or $8$ hidden layers of $256$ processing elements. The post-activation function used was the hyperbolic tangent and the top layer was chosen to be a maximum-entropy classifier (i.e., a softmax function). The output layer objective for all algorithms was to minimize the categorical negative log likelihood.

Parameters were initialized using a scheme that gave best performance on the validation split of each dataset on a per-algorithm basis. Though we wanted to use very simple initialization schemes for all algorithms, in preliminary experiments, we found that the feedback alignment algorithms as well as \emph{DTP} (and \emph{DTP}-$\sigma$) worked best when using  a uniform fan-in-fan-out scheme \cite{glorot2010understanding}. \cite{ororbia2018conducting} confirms this result, originally showing how these algorithms often are unstable or fail to perform well using initializations based on simple uniform or Gaussian distributions. For LRA-E, however, we initialized the parameters using a zero-mean Gaussian distribution (variance of $0.05$).

\begin{table*}[!t]
\centering
\begin{tabular}{l|r|r||r|r||r|r}
\multicolumn{1}{l}{}&\multicolumn{2}{|c}{\begin{tabular}[x]{@{}c@{}}\textbf{3 Layers}\\\end{tabular}}&\multicolumn{2}{|c}{\begin{tabular}[x]{@{}c@{}}\textbf{5 Layers}\\\end{tabular}}&\multicolumn{2}{|c}{\begin{tabular}[x]{@{}c@{}}\textbf{8 Layers}\\\end{tabular}}\tabularnewline
\multicolumn{1}{l}{\textbf{Model}}&\multicolumn{1}{|c}{\begin{tabular}[x]{@{}c@{}}\textbf{Train}\\\end{tabular}}&\multicolumn{1}{|c}{\begin{tabular}[x]{@{}c@{}}\textbf{Test}\\\end{tabular}}&\multicolumn{1}{|c}{\begin{tabular}[x]{@{}c@{}}\textbf{Train}\\\end{tabular}}&\multicolumn{1}{|c}{\begin{tabular}[x]{@{}c@{}}\textbf{Test}\\\end{tabular}}&\multicolumn{1}{|c}{\begin{tabular}[x]{@{}c@{}}\textbf{Train}\\\end{tabular}}&\multicolumn{1}{|c}{\begin{tabular}[x]{@{}c@{}}\textbf{Test}\\\end{tabular}}\tabularnewline
\hline
\textit{Backprop} & $1.78$ & $3.02$ & $2.4$ & $2.98$ & $2.91$ & $3.02$\tabularnewline
\textit{Equil-Prop} & $3.82$ & $4.99$ & $7.59$ & $9.21$ & $89.96$ & $90.91$\tabularnewline
\textit{RFA} & $3.01$ & $3.13$ & $2.99$ & $3.4$ & $3.59$ & $3.76$\tabularnewline
\textit{DFA} & $4.07$ & $4.17$ & $3.71$ & $3.88$ & $3.81$ & $3.85$\tabularnewline
\textit{DTP} & $0.74$ & $2.8$ & $4.408$ & $4.94$ & $10.89$ & $10.1$\tabularnewline
\textit{DTP}-$\sigma$ (ours) & $0.00$ & $2.38$ & $0.00$ & $2.57$ & $0.00$ & $2.56$\tabularnewline
\textit{LRA}-E (ours) & $0.86$ & $2.20$ & $0.16$ & $1.97$ & $0.08$ & $2.55$\tabularnewline
\hline
\end{tabular}
\vspace{-0.1cm}
\caption{MNIST supervised classification results.}
\label{mnist_results}
\vspace{0.2cm}
\centering
\begin{tabular}{l|r|r||r|r||r|r}
\multicolumn{1}{l}{}&\multicolumn{2}{|c}{\begin{tabular}[x]{@{}c@{}}\textbf{3 Layers}\\\end{tabular}}&\multicolumn{2}{|c}{\begin{tabular}[x]{@{}c@{}}\textbf{5 Layers}\\\end{tabular}}&\multicolumn{2}{|c}{\begin{tabular}[x]{@{}c@{}}\textbf{8 Layers}\\\end{tabular}}\tabularnewline
\multicolumn{1}{l}{\textbf{Model}}&\multicolumn{1}{|c}{\begin{tabular}[x]{@{}c@{}}\textbf{Train}\\\end{tabular}}&\multicolumn{1}{|c}{\begin{tabular}[x]{@{}c@{}}\textbf{Test}\\\end{tabular}}&\multicolumn{1}{|c}{\begin{tabular}[x]{@{}c@{}}\textbf{Train}\\\end{tabular}}&\multicolumn{1}{|c}{\begin{tabular}[x]{@{}c@{}}\textbf{Test}\\\end{tabular}}&\multicolumn{1}{|c}{\begin{tabular}[x]{@{}c@{}}\textbf{Train}\\\end{tabular}}&\multicolumn{1}{|c}{\begin{tabular}[x]{@{}c@{}}\textbf{Test}\\\end{tabular}}\tabularnewline
\hline
\textit{Backprop} & $12.08$ & $14.89$ & $12.1$ & $12.98$ & $11.55$ & $13.21$\tabularnewline
\textit{Equil-Prop} & $14.72$ & $14.01$ & $16.56$ & $20.97$ & $90.12$ & $89.78$\tabularnewline
\textit{RFA} & $11.99$ & $12.74$ & $12.09$ & $12.89$ & $12.03$ & $12.71$\tabularnewline
\textit{DFA} & $13.04$ & $13.41$ & $12.58$ & $13.09$ & $11.59$ & $13.01$\tabularnewline
\textit{DTP} & $13.6$ & $15.03$ & $21.078$ & $19.66$ & $21.838$ & $17.58$\tabularnewline
\textit{DTP}-$\sigma$ (ours) & $7.5$ & $13.95$ & $6.34$ & $12.99$ & $6.5$ & $13.01$\tabularnewline
\textit{LRA}-E (ours) & $11.25$ & $13.51$ & $9.84$ & $12.31$ & $9.74$ & $12.69$ \tabularnewline
\hline
\end{tabular}
\vspace{-0.1cm}
\caption{Fashion MNIST supervised classification results.}
\label{fmnist_results}
\end{table*}

The choice of parameter update rule was also somewhat dependent on the learning algorithm employed. Again, as shown in \cite{ororbia2018conducting}, it is difficult to get good, stable performance from algorithms, such as the original DTP, when using simple SGD. As done in \cite{lee2015targetprop}, we used the RMSprop \cite{tieleman2012rmsprop} adaptive learning rate with a global step size of $\lambda = 0.001$. For Backprop, RFA, DFA, and LRA-E, we were able to use SGD ($\lambda = 0.01$).

\subsection{Classification Performance}
\label{sec:supervised}

\begin{table*}[!t]
\centering
\begin{tabular}{l|r|r||r|r||r|r}
\multicolumn{1}{l}{\textbf{}}&\multicolumn{1}{|c}{\begin{tabular}[x]{@{}c@{}}\textbf{SGD}\\\end{tabular}}&\multicolumn{1}{c}{\begin{tabular}[x]{@{}c@{}}\textbf{}\\\end{tabular}}&\multicolumn{1}{|c}{\begin{tabular}[x]{@{}c@{}}\textbf{Adam}\\\end{tabular}}&\multicolumn{1}{c}{\begin{tabular}[x]{@{}c@{}}\textbf{}\\\end{tabular}}&\multicolumn{1}{|c}{\begin{tabular}[x]{@{}c@{}}\textbf{RMSprop}\\\end{tabular}}&\multicolumn{1}{c}{\begin{tabular}[x]{@{}c@{}}\textbf{}\\\end{tabular}}\tabularnewline
\multicolumn{1}{l}{\textbf{Model}}&\multicolumn{1}{|c}{\begin{tabular}[x]{@{}c@{}}\textbf{Train}\\\end{tabular}}&\multicolumn{1}{|c}{\begin{tabular}[x]{@{}c@{}}\textbf{Test}\\\end{tabular}}&\multicolumn{1}{|c}{\begin{tabular}[x]{@{}c@{}}\textbf{Train}\\\end{tabular}}&\multicolumn{1}{|c}{\begin{tabular}[x]{@{}c@{}}\textbf{Test}\\\end{tabular}}&\multicolumn{1}{|c}{\begin{tabular}[x]{@{}c@{}}\textbf{Train}\\\end{tabular}}&\multicolumn{1}{|c}{\begin{tabular}[x]{@{}c@{}}\textbf{Test}\\\end{tabular}}\tabularnewline
\hline
\textit{LRA, MNIST} & $0.86$ & $2.20$ & $0.00$ & $1.75$ & $0.69$ & $2.02$ \tabularnewline
\textit{LRA, Fashion MNIST} & $11.25$ & $13.51$ & $5.38$ & $12.42$ & $12.67$ & $14.90$ \tabularnewline
\hline
\end{tabular}
\caption{Effect of the update rule on LRA when training a 3-layer MLP on MNIST.}
\label{optimization_results}
\end{table*}

\begin{figure*}[!t]
\centering     
\subfigure[DFA.]{\label{fig:dfa}\includegraphics[width=33mm]{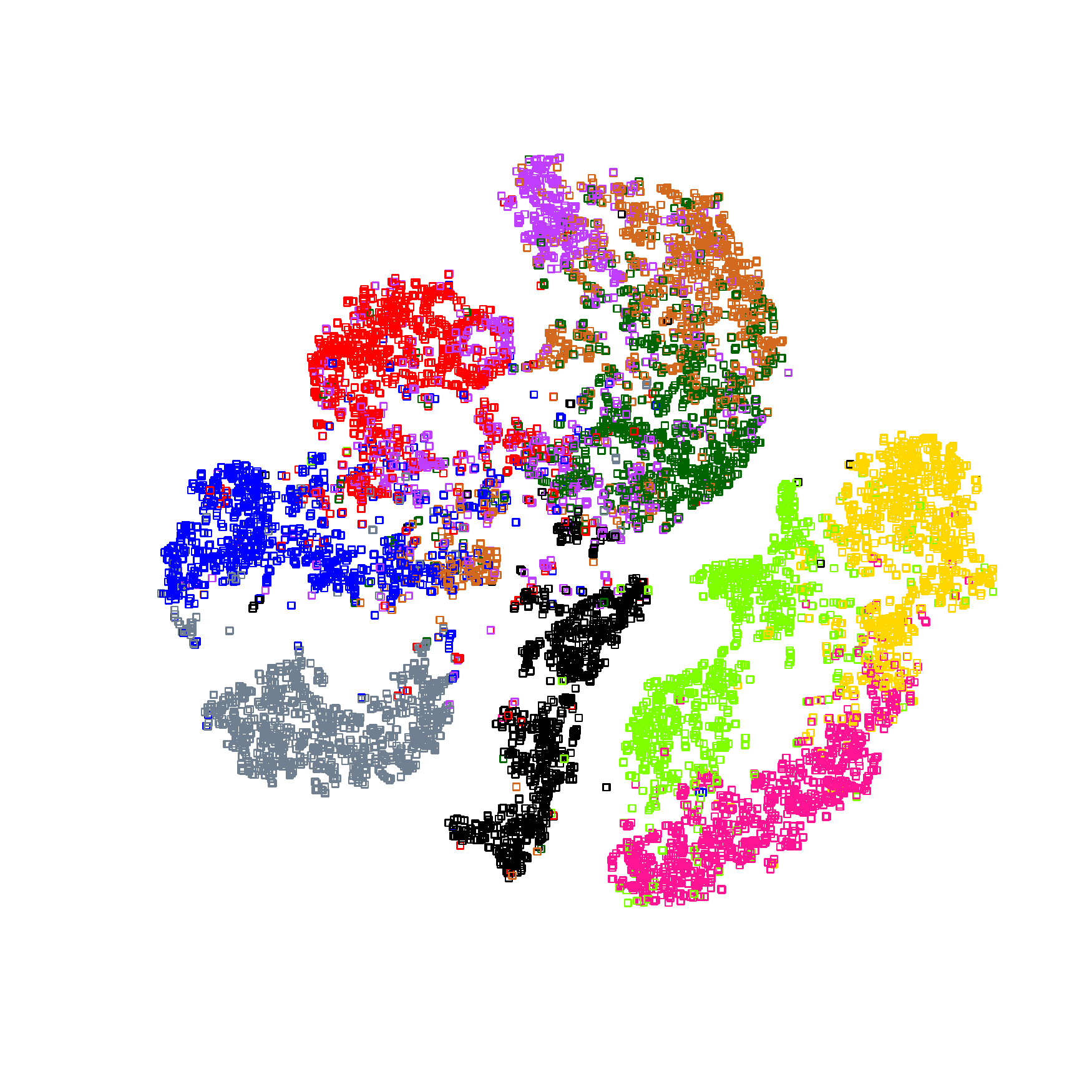}
}
\subfigure[Equil-Prop.]{\label{fig:eqprop}\includegraphics[width=33mm]{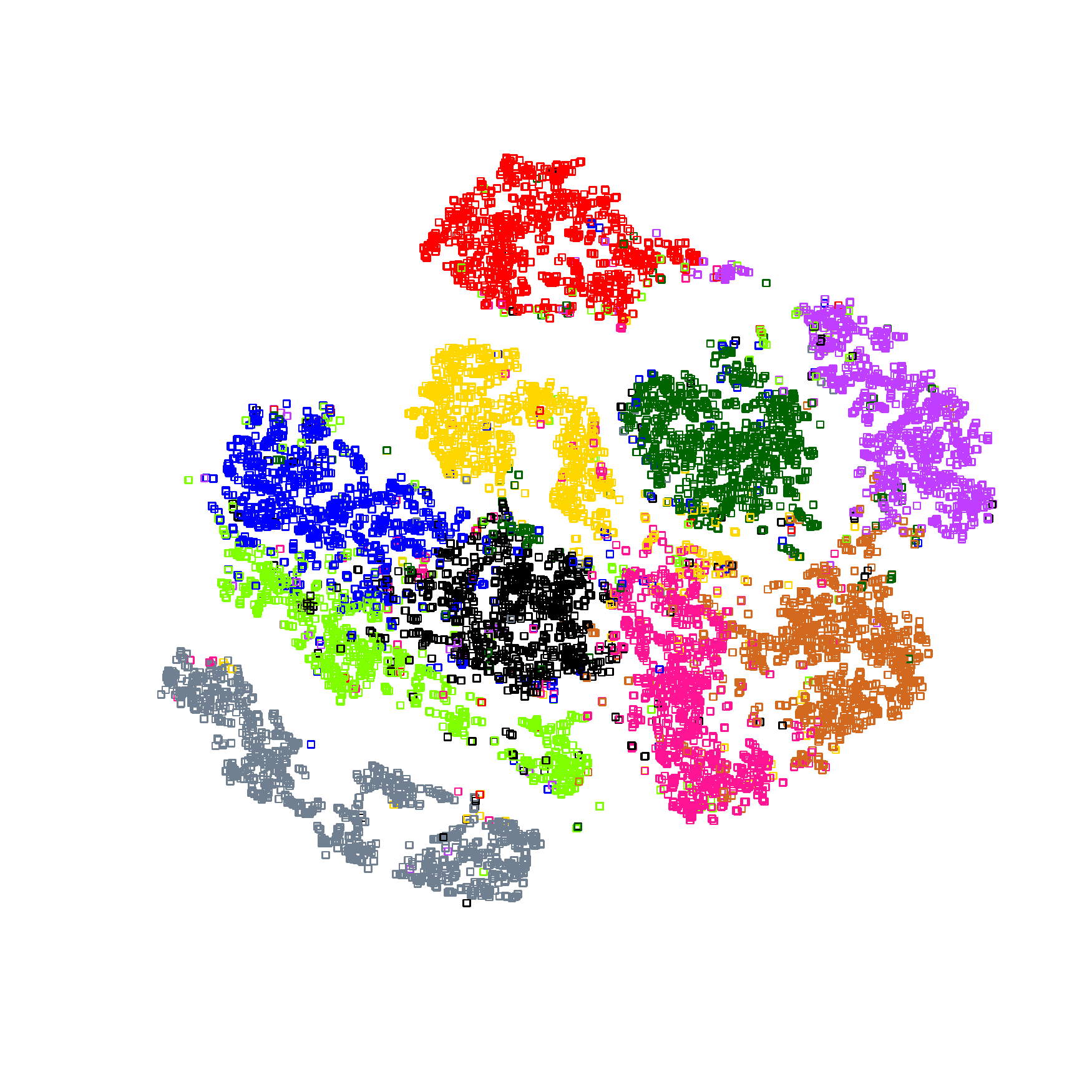}
}
\subfigure[DTP-$\sigma$.]{\label{fig:dtp}\includegraphics[width=33mm]{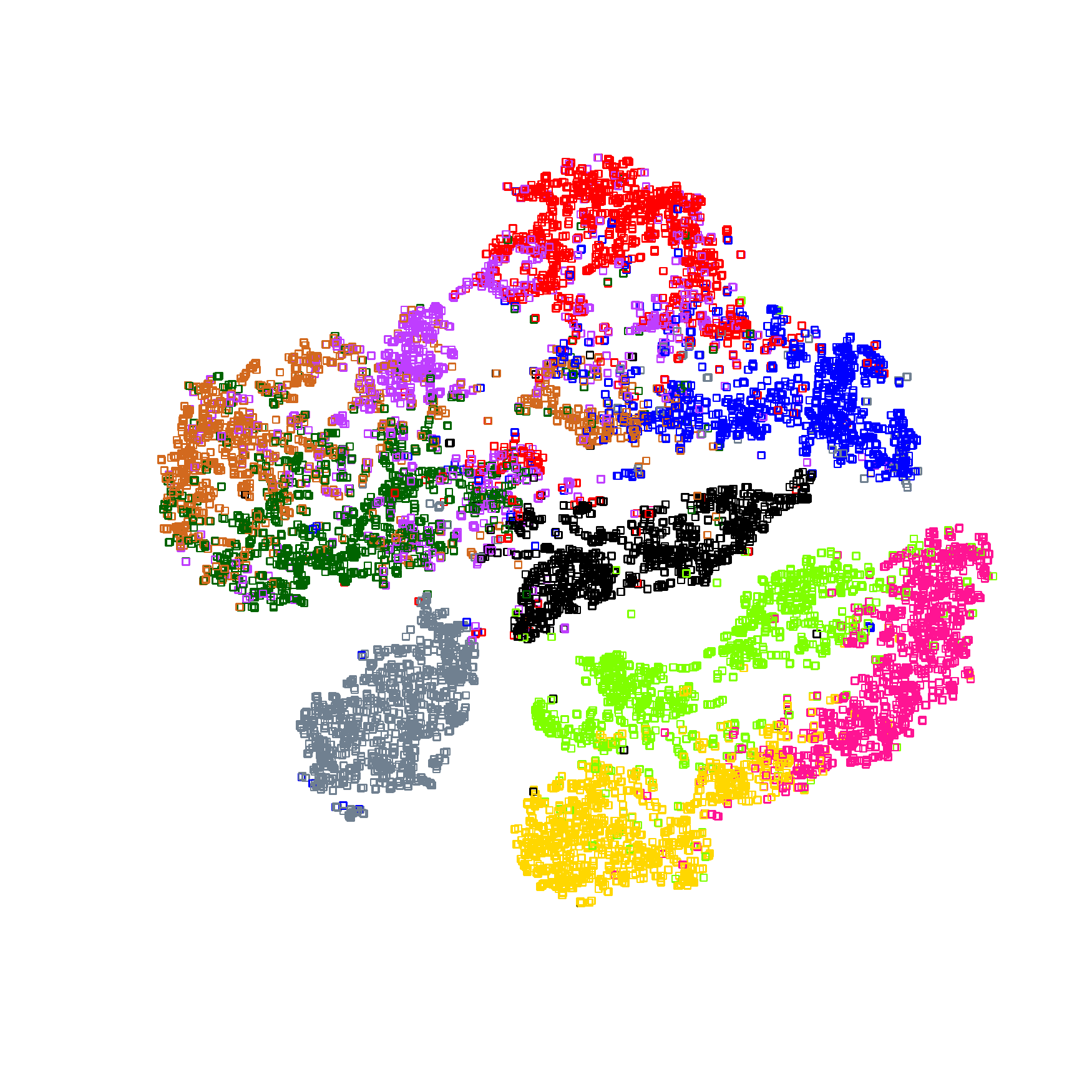}
}
\subfigure[LRA-E.]{\label{fig:lra_e}\includegraphics[width=33mm]{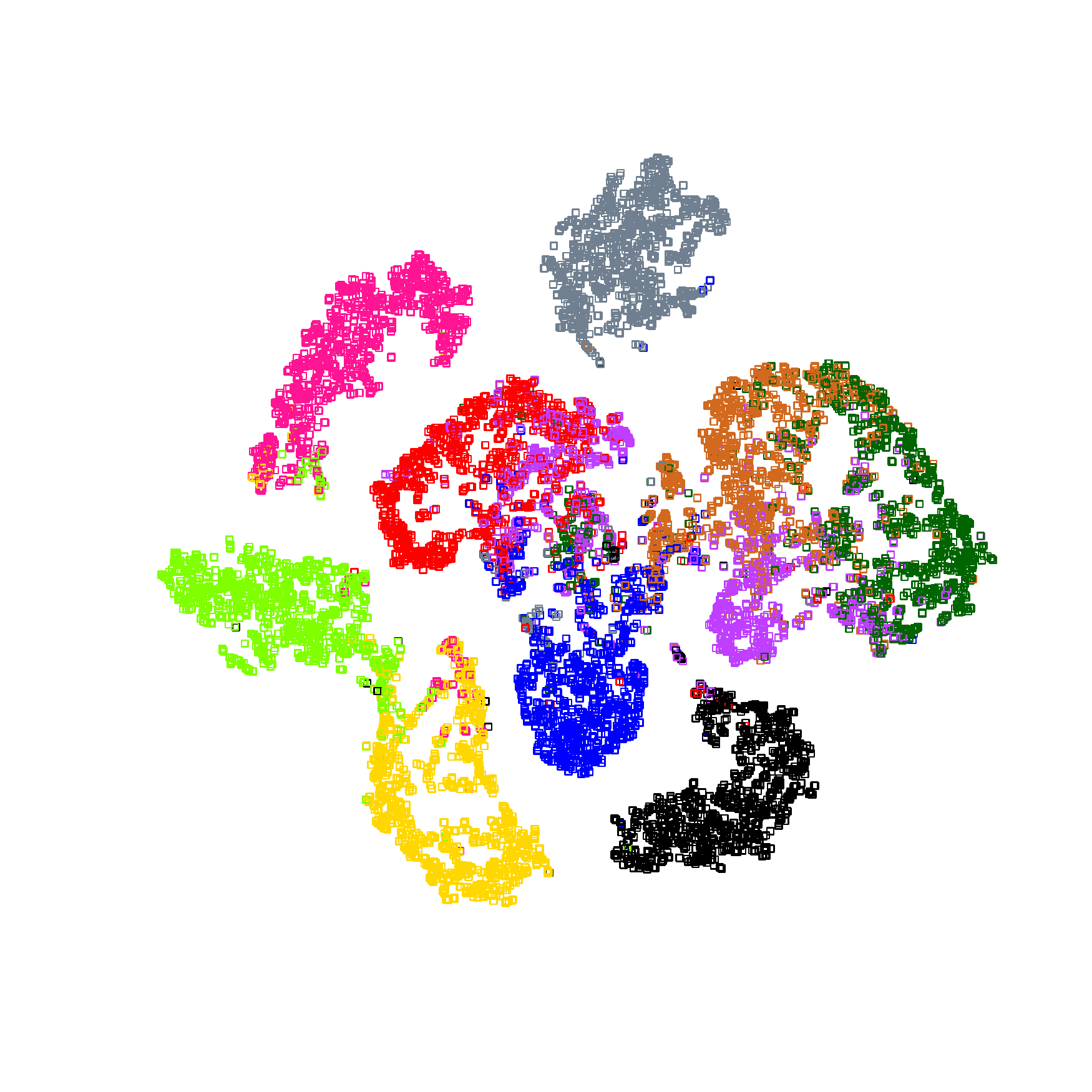}
}
\caption{Visualization of the topmost hidden layer of a 5-layer MLP trained by DFA, Equil-Prop, DTP-$\sigma$, and LRA-E.}
\label{fig:tsne_plots}
\end{figure*}

In this experiment, we compare all of the algorithms discussed earlier. These include back-propagation (Backprop), Random Feedback Alignment (RFA) \cite{lillicrap2014random}, Direct Feedback Alignment (DFA) \cite{nokland2016direct}, Equilibrium Propagation \cite{scellier2017equilibrium} (Equil-Prop)
and the original Difference Target Propagation \cite{lee2015difference} (DTP). Our algorithms include our proposed, improved version of DTP (\textit{DTP}-$\sigma$) and the proposed error-driven Local Representation Alignment (LRA-E). 

The results of our experiments are presented in Tables \ref{mnist_results} and \ref{fmnist_results}. Test and training scores are reported for the set of model parameters that had lowest validation error. Observe that LRA-E is the most stable and consistently well-performing algorithm compared to the other backprop alternatives, closely followed by our \emph{DTP}-$\sigma$. More importantly, algorithms like Equil-Prop and DTP appear to break down when training deeper networks, i.e., the 8-layer MLP. Note that while DTP was used to successfully train a 7-layer network of 240 units (using RMSprop) \cite{lee2015difference}, we followed the same settings reported for networks deeper than $7$ and in our experiments uncovered that the algorithm begins to struggle as the layers are made wider, starting with the width of $256$. However, this problem is rectified using DTP-$\sigma$, leading to much more stable performance and even to cases where the algorithm completely overfits the training set (as in the case of 3 and 5 layers for MNIST). Nonetheless, LRA-E still performs best with respect to generalization across both datasets, despite using a na\"ive initialization scheme. Table \ref{optimization_results} shows the results of using update rules other than SGD for LRA-E, e.g., Adam \cite{kingma2014adam} or RMSprop \cite{tieleman2012rmsprop} for a 3-layer MLP, (global step size $0.001$ for both algorithms). We see that LRA-E is compatible with other learning rate schemes and reaches better generalization performance when using them.

\begin{figure*}[!t]
\centering     
\subfigure[5-layer MLP.]{\label{fig:5mlp}\includegraphics[width=73mm]{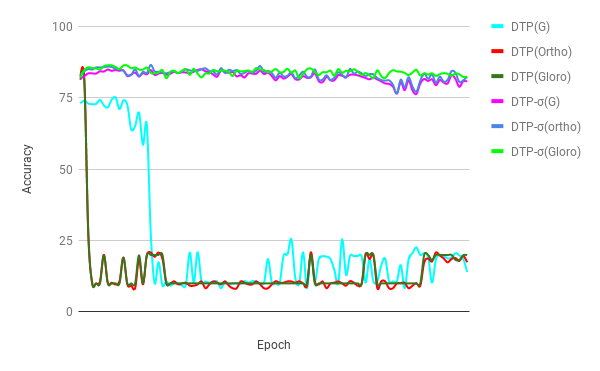}
}
\subfigure[8-layer MLP]{\label{fig:8mlp}\includegraphics[width=73mm]{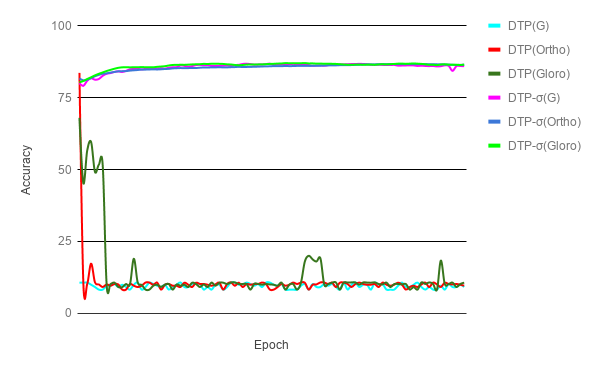}
}
\caption{Validation accuracy of \emph{DTP} vs our proposed \emph{DTP}-$\sigma$, as a function of epoch.}
\label{fig:dtp_plots}
\end{figure*}

Figure \ref{fig:tsne_plots} displays a t-SNE \cite{maaten2008visualizing} visualization of the top-most hidden layer of a learned 5-layer MLP using either DFA, Equil-Prop, \emph{DTP}-$\sigma$, and LRA-E on Fashion MNIST samples. Qualitatively, we see that all learning algorithms extract representations that separate out the data points reasonably well, at least in the sense that points are clustered based on clothing type. However, it appears that \emph{LRA-E} representations yield more strongly separated clusters, as evidenced by somewhat wider gaps between them, especially around the pink, blue, and black colored clusters.

Finally, DTP, as also mentioned in \cite{ororbia2018conducting}, appears to be quite sensitive to its initialization scheme. For both MNIST and Fashion MNIST, we trained DTP and our proposed \emph{DTP}-$\sigma$ with three different settings, including random orthogonal (Ortho), fan-in-fan-out (Gloro), and simple zero-mean Gaussian (G) initializations. Figure \ref{fig:dtp_plots} shows the validation accuracy curves of DTP and \textit{DTP}-$\sigma$ as a function of epoch for 5 and 8 layer networks 
with various weight initializations. As shown in Figure \ref{fig:dtp_plots}, \emph{DTP} is highly unstable as the network gets deeper while \emph{DTP}-$\sigma$ is not. Furthermore, \emph{DTP}-$\sigma$'s performance appears to be less dependent on the weight initialization scheme.
Thus, our experiments show promising evidence of \emph{DTP}-$\sigma$'s generalization improvement over the original \emph{DTP}. Moreso, as indicated by Tables \ref{mnist_results} and \ref{fmnist_results}, \emph{DTP}-$\sigma$ can, overall, perform nearly as well as \emph{LRA-E}.

\section{Conclusions}
\label{conc}
In this paper, we proposed two learning algorithms: error-driven Local Representation Alignment and adaptive noise Difference Target Propagation. On two classification benchmarks, we show strong positive results when training deep multilayer perceptrons. 
With respect to other types of neural structures, e.g., locally connected ones, we would expect our proposed algorithms to work well, especially LRA-E, since the target computation/error unit mechanism is agnostic to the underlying building blocks of the feedforward model (permitting extension to models such as residual networks).
Future work will include adapting these algorithms to larger-scale tasks requiring more complex, exotic architectures.

%


\bibliographystyle{aaai}
\bibliography{lra_recirc}

\begin{thebibliography}{}

\bibitem[\protect\citeauthoryear{Balduzzi, Vanchinathan, and
  Buhmann}{2015}]{balduzzi2015kickback}
Balduzzi, D.; Vanchinathan, H.; and Buhmann, J.~M.
\newblock 2015.
\newblock Kickback cuts backprop's red-tape: Biologically plausible credit
  assignment in neural networks.
\newblock In {\em AAAI},  485--491.

\bibitem[\protect\citeauthoryear{Bengio \bgroup et al\mbox.\egroup
  }{2007}]{bengio_greedy_2007}
Bengio, Y.; Lamblin, P.; Popovici, D.; Larochelle, H.; et~al.
\newblock 2007.
\newblock Greedy layer-wise training of deep networks.
\newblock {\em Advances in Neural Information Processing Systems} 19:153.

\bibitem[\protect\citeauthoryear{Bengio \bgroup et al\mbox.\egroup
  }{2015}]{bengio2015towards}
Bengio, Y.; Lee, D.-H.; Bornschein, J.; Mesnard, T.; and Lin, Z.
\newblock 2015.
\newblock Towards biologically plausible deep learning.
\newblock {\em arXiv preprint arXiv:1502.04156}.

\bibitem[\protect\citeauthoryear{Cordo \bgroup et al\mbox.\egroup
  }{1996}]{Cordo}
Cordo, P.; Inglis, J.~T.; Verschueren, S.; Collins, J.~J.; Merfeld, D.~M.;
  Rosenblum, S.; Buckley, S.; and Moss, F.
\newblock 1996.
\newblock {{N}oise in human muscle spindles}.
\newblock {\em Nature} 383(6603):769--770.

\bibitem[\protect\citeauthoryear{D. and Yngve}{}]{adrian}
D., A.~E., and Yngve, Z.
\newblock The impulses produced by sensory nerve‐endings.
\newblock {\em The Journal of Physiology} 61(2):151--171.

\bibitem[\protect\citeauthoryear{Faisal, Selen, and Wolpert}{2008}]{faizal}
Faisal, A.~A.; Selen, L.~P.; and Wolpert, D.~M.
\newblock 2008.
\newblock {{N}oise in the nervous system}.
\newblock {\em Nat. Rev. Neurosci.} 9(4):292--303.

\bibitem[\protect\citeauthoryear{Feng, Yu, and Zhou}{2018}]{feng2018multi}
Feng, J.; Yu, Y.; and Zhou, Z.-H.
\newblock 2018.
\newblock Multi-layered gradient boosting decision trees.
\newblock {\em arXiv preprint arXiv:1806.00007}.

\bibitem[\protect\citeauthoryear{Glorot and
  Bengio}{2010}]{glorot2010understanding}
Glorot, X., and Bengio, Y.
\newblock 2010.
\newblock Understanding the difficulty of training deep feedforward neural
  networks.
\newblock In {\em Proceedings of the Thirteenth International Conference on
  Artificial Intelligence and Statistics},  249--256.

\bibitem[\protect\citeauthoryear{Grossberg}{1982}]{grossberg1982does}
Grossberg, S.
\newblock 1982.
\newblock How does a brain build a cognitive code?
\newblock In {\em Studies of mind and brain}. Springer.
\newblock  1--52.

\bibitem[\protect\citeauthoryear{Grossberg}{1987}]{grossberg_resonance_1987}
Grossberg, S.
\newblock 1987.
\newblock Competitive learning: From interactive activation to adaptive
  resonance.
\newblock {\em Cognitive Science} 11(1):23 -- 63.

\bibitem[\protect\citeauthoryear{Hinton and
  McClelland}{1988}]{hinton1988learning}
Hinton, G.~E., and McClelland, J.~L.
\newblock 1988.
\newblock Learning representations by recirculation.
\newblock In {\em Neural information processing systems},  358--366.

\bibitem[\protect\citeauthoryear{Huang and Rao}{2011}]{huang2011predictive}
Huang, Y., and Rao, R.~P.
\newblock 2011.
\newblock Predictive coding.
\newblock {\em Wiley Interdisciplinary Reviews: Cognitive Science}
  2(5):580--593.

\bibitem[\protect\citeauthoryear{Jaderberg \bgroup et al\mbox.\egroup
  }{2016}]{jaderberg2016decoupled}
Jaderberg, M.; Czarnecki, W.~M.; Osindero, S.; Vinyals, O.; Graves, A.; and
  Kavukcuoglu, K.
\newblock 2016.
\newblock Decoupled neural interfaces using synthetic gradients.
\newblock {\em arXiv preprint arXiv:1608.05343}.

\bibitem[\protect\citeauthoryear{Kavukcuoglu, Ranzato, and
  LeCun}{2010}]{kavukcuoglu2010fast}
Kavukcuoglu, K.; Ranzato, M.; and LeCun, Y.
\newblock 2010.
\newblock Fast inference in sparse coding algorithms with applications to
  object recognition.
\newblock {\em arXiv preprint arXiv:1010.3467}.

\bibitem[\protect\citeauthoryear{Kingma and Ba}{2014}]{kingma2014adam}
Kingma, D., and Ba, J.
\newblock 2014.
\newblock Adam: A method for stochastic optimization.
\newblock {\em arXiv preprint arXiv:1412.6980}.

\bibitem[\protect\citeauthoryear{Kruglikov and Dertinger}{1994}]{Kruglikov}
Kruglikov, I.~L., and Dertinger, H.
\newblock 1994.
\newblock {{S}tochastic resonance as a possible mechanism of amplification of
  weak electric signals in living cells}.
\newblock {\em Bioelectromagnetics} 15(6):539--547.

\bibitem[\protect\citeauthoryear{Laughlin, de~Ruyter~van Steveninck, and
  Anderson}{1998}]{Laughlin}
Laughlin, S.~B.; de~Ruyter~van Steveninck, R.~R.; and Anderson, J.~C.
\newblock 1998.
\newblock {{T}he metabolic cost of neural information}.
\newblock {\em Nat. Neurosci.} 1(1):36--41.

\bibitem[\protect\citeauthoryear{Lee \bgroup et al\mbox.\egroup
  }{2014}]{lee_deeply-supervised_2014}
Lee, C.-Y.; Xie, S.; Gallagher, P.; Zhang, Z.; and Tu, Z.
\newblock 2014.
\newblock {Deeply-Supervised Nets}.
\newblock {\em {arXiv}:1409.5185 [cs, stat]}.

\bibitem[\protect\citeauthoryear{Lee \bgroup et al\mbox.\egroup
  }{2015a}]{lee2015difference}
Lee, D.-H.; Zhang, S.; Fischer, A.; and Bengio, Y.
\newblock 2015a.
\newblock Difference target propagation.
\newblock In {\em Joint European Conference on Machine Learning and Knowledge
  Discovery in Databases},  498--515.
\newblock Springer.

\bibitem[\protect\citeauthoryear{Lee \bgroup et al\mbox.\egroup
  }{2015b}]{lee2015targetprop}
Lee, D.-H.; Zhang, S.; Fischer, A.; and Bengio, Y.
\newblock 2015b.
\newblock Difference target propagation.
\newblock In {\em Proceedings of the 2015th European Conference on Machine
  Learning and Knowledge Discovery in Databases - Volume Part I}, ECMLPKDD'15,
  498--515.
\newblock Switzerland: Springer.

\bibitem[\protect\citeauthoryear{Li and Liu}{2018}]{li2018predictive}
Li, J., and Liu, H.
\newblock 2018.
\newblock Predictive coding machine for compressed sensing and image denoising.
\newblock In {\em AAAI}.

\bibitem[\protect\citeauthoryear{Liao, Leibo, and
  Poggio}{2016}]{liao2016important}
Liao, Q.; Leibo, J.~Z.; and Poggio, T.~A.
\newblock 2016.
\newblock How important is weight symmetry in backpropagation?
\newblock In {\em AAAI},  1837--1844.

\bibitem[\protect\citeauthoryear{Lillicrap \bgroup et al\mbox.\egroup
  }{2014}]{lillicrap2014random}
Lillicrap, T.~P.; Cownden, D.; Tweed, D.~B.; and Akerman, C.~J.
\newblock 2014.
\newblock Random feedback weights support learning in deep neural networks.
\newblock {\em arXiv preprint arXiv:1411.0247}.

\bibitem[\protect\citeauthoryear{Lillicrap \bgroup et al\mbox.\egroup
  }{2016}]{lillicrap2016random}
Lillicrap, T.~P.; Cownden, D.; Tweed, D.~B.; and Akerman, C.~J.
\newblock 2016.
\newblock Random synaptic feedback weights support error backpropagation for
  deep learning.
\newblock {\em Nature communications} 7:13276.

\bibitem[\protect\citeauthoryear{Maaten and
  Hinton}{2008}]{maaten2008visualizing}
Maaten, L. v.~d., and Hinton, G.
\newblock 2008.
\newblock Visualizing data using t-sne.
\newblock {\em Journal of machine learning research} 9(Nov):2579--2605.

\bibitem[\protect\citeauthoryear{N{\o}kland}{2016}]{nokland2016direct}
N{\o}kland, A.
\newblock 2016.
\newblock Direct feedback alignment provides learning in deep neural networks.
\newblock In {\em Advances in Neural Information Processing Systems},
  1037--1045.

\bibitem[\protect\citeauthoryear{Olshausen and
  Field}{1997}]{olshausen1997sparse}
Olshausen, B.~A., and Field, D.~J.
\newblock 1997.
\newblock Sparse coding with an overcomplete basis set: A strategy employed by
  v1?
\newblock {\em Vision research} 37(23):3311--3325.

\bibitem[\protect\citeauthoryear{O'Reilly}{1996}]{o1996biologically}
O'Reilly, R.~C.
\newblock 1996.
\newblock Biologically plausible error-driven learning using local activation
  differences: The generalized recirculation algorithm.
\newblock {\em Neural computation} 8(5):895--938.

\bibitem[\protect\citeauthoryear{Ororbia \bgroup et al\mbox.\egroup
  }{2018}]{ororbia2018conducting}
Ororbia, A.~G.; Mali, A.; Kifer, D.; and Giles, C.~L.
\newblock 2018.
\newblock Conducting credit assignment by aligning local representations.
\newblock {\em arXiv preprint arXiv:1803.01834}.

\bibitem[\protect\citeauthoryear{Ororbia~II \bgroup et al\mbox.\egroup
  }{2015}]{ororbia_deep_hybrid_2015a}
Ororbia~II, A.~G.; Reitter, D.; Wu, J.; and Giles, C.~L.
\newblock 2015.
\newblock Online learning of deep hybrid architectures for semi-supervised
  categorization.
\newblock In {\em Machine {Learning} and {Knowledge} {Discovery} in {Databases}
  ({Proceedings}, {ECML} {PKDD} 2015)}, volume 9284 of {\em Lecture {Notes} in
  {Computer} {Science}}. Porto, Portugal: Springer.
\newblock  516--532.

\bibitem[\protect\citeauthoryear{Ororbia~II \bgroup et al\mbox.\egroup
  }{2017}]{ororbia2017learning}
Ororbia~II, A.~G.; Haffner, P.; Reitter, D.; and Giles, C.~L.
\newblock 2017.
\newblock Learning to adapt by minimizing discrepancy.
\newblock {\em arXiv preprint arXiv:1711.11542}.

\bibitem[\protect\citeauthoryear{Ororbia~II, Giles, and
  Reitter}{2015}]{ororbia2015online}
Ororbia~II, A.~G.; Giles, C.~L.; and Reitter, D.
\newblock 2015.
\newblock Online semi-supervised learning with deep hybrid boltzmann machines
  and denoising autoencoders.
\newblock {\em arXiv preprint arXiv:1511.06964}.

\bibitem[\protect\citeauthoryear{Ororbia~II, Kifer, and
  Giles}{2017}]{ororbia2017unifying}
Ororbia~II, A.~G.; Kifer, D.; and Giles, C.~L.
\newblock 2017.
\newblock Unifying adversarial training algorithms with data gradient
  regularization.
\newblock {\em Neural computation} 29(4):867--887.

\bibitem[\protect\citeauthoryear{Panichello, Cheung, and
  Bar}{2013}]{panichello2013predictive}
Panichello, M.; Cheung, O.; and Bar, M.
\newblock 2013.
\newblock Predictive feedback and conscious visual experience.
\newblock {\em Frontiers in Psychology} 3:620.

\bibitem[\protect\citeauthoryear{Pascanu, Mikolov, and
  Bengio}{2013}]{pascanu2013difficulty}
Pascanu, R.; Mikolov, T.; and Bengio, Y.
\newblock 2013.
\newblock On the difficulty of training recurrent neural networks.
\newblock In {\em International Conference on Machine Learning},  1310--1318.

\bibitem[\protect\citeauthoryear{Rao and Ballard}{1997}]{rao1997dynamic}
Rao, R.~P., and Ballard, D.~H.
\newblock 1997.
\newblock Dynamic model of visual recognition predicts neural response
  properties in the visual cortex.
\newblock {\em Neural computation} 9(4):721--763.

\bibitem[\protect\citeauthoryear{Rao and Ballard}{1999}]{rao1999predictive}
Rao, R.~P., and Ballard, D.~H.
\newblock 1999.
\newblock Predictive coding in the visual cortex: a functional interpretation
  of some extra-classical receptive-field effects.
\newblock {\em Nature neuroscience} 2(1).

\bibitem[\protect\citeauthoryear{Rumelhart, Hinton, and
  Williams}{1988}]{rumelhart1988backprop}
Rumelhart, D.~E.; Hinton, G.~E.; and Williams, R.~J.
\newblock 1988.
\newblock Neurocomputing: Foundations of research.
\newblock Cambridge, MA, USA: MIT Press.
\newblock chapter Learning Representations by Back-propagating Errors,
  696--699.

\bibitem[\protect\citeauthoryear{Sarpeshkar}{1998}]{Sarpeshkar}
Sarpeshkar, R.
\newblock 1998.
\newblock {{A}nalog versus digital: extrapolating from electronics to
  neurobiology}.
\newblock {\em Neural Comput} 10(7):1601--1638.

\bibitem[\protect\citeauthoryear{Scellier and
  Bengio}{2017}]{scellier2017equilibrium}
Scellier, B., and Bengio, Y.
\newblock 2017.
\newblock Equilibrium propagation: Bridging the gap between energy-based models
  and backpropagation.
\newblock {\em Frontiers in computational neuroscience} 11:24.

\bibitem[\protect\citeauthoryear{Shadlen and Newsome}{1998}]{pmid9570816}
Shadlen, M.~N., and Newsome, W.~T.
\newblock 1998.
\newblock {{T}he variable discharge of cortical neurons: implications for
  connectivity, computation, and information coding}.
\newblock {\em J. Neurosci.} 18(10):3870--3896.

\bibitem[\protect\citeauthoryear{Shu \bgroup et al\mbox.\egroup }{2003}]{shu}
Shu, Y.; Hasenstaub, A.; Badoual, M.; Bal, T.; and McCormick, D.~A.
\newblock 2003.
\newblock {{B}arrages of synaptic activity control the gain and sensitivity of
  cortical neurons}.
\newblock {\em J. Neurosci.} 23(32):10388--10401.

\bibitem[\protect\citeauthoryear{Srivastava \bgroup et al\mbox.\egroup
  }{2014}]{srivastava2014dropout}
Srivastava, N.; Hinton, G.; Krizhevsky, A.; Sutskever, I.; and Salakhutdinov,
  R.
\newblock 2014.
\newblock Dropout: A simple way to prevent neural networks from overfitting.
\newblock {\em The Journal of Machine Learning Research} 15(1):1929--1958.

\bibitem[\protect\citeauthoryear{Tieleman and
  Hinton}{2012}]{tieleman2012rmsprop}
Tieleman, T., and Hinton, G.
\newblock 2012.
\newblock {Lecture 6.5---RmsProp: Divide the gradient by a running average of
  its recent magnitude}.
\newblock COURSERA: Neural Networks for Machine Learning.

\bibitem[\protect\citeauthoryear{Tolhurst, Movshon, and
  Dean}{1983}]{TOLHURST1983775}
Tolhurst, D.; Movshon, J.; and Dean, A.
\newblock 1983.
\newblock The statistical reliability of signals in single neurons in cat and
  monkey visual cortex.
\newblock {\em Vision Research} 23(8):775 -- 785.

\bibitem[\protect\citeauthoryear{Tomko and Crapper}{1974}]{TOMKO1974405}
Tomko, G.~J., and Crapper, D.~R.
\newblock 1974.
\newblock Neuronal variability: non-stationary responses to identical visual
  stimuli.
\newblock {\em Brain Research} 79(3):405 -- 418.

\bibitem[\protect\citeauthoryear{Whittington and
  Bogacz}{2017}]{whittington2017equivalence}
Whittington, J. C.~R., and Bogacz, R.
\newblock 2017.
\newblock An approximation of the error backpropagation algorithm in a
  predictive coding network with local hebbian synaptic plasticity.
\newblock {\em Neural Computation} 29(5):1229--1262.
\newblock PMID: 28333583.

\bibitem[\protect\citeauthoryear{Xiao, Rasul, and
  Vollgraf}{2017}]{xiao2017fashion}
Xiao, H.; Rasul, K.; and Vollgraf, R.
\newblock 2017.
\newblock Fashion-mnist: a novel image dataset for benchmarking machine
  learning algorithms.
\newblock {\em arXiv preprint arXiv:1708.07747}.

\end{thebibliography}

\end{document}